\documentclass[conference]{IEEEtran}
\IEEEoverridecommandlockouts
\usepackage{times}
\usepackage{epsfig}
\usepackage{graphicx}
\usepackage{subfigure}
\usepackage{multirow}
\usepackage{multicol}
\usepackage{cite}
\usepackage{amsmath,amssymb,amsfonts}
\usepackage{textcomp}
\usepackage{xcolor}
\usepackage{algorithm}  
\usepackage{algorithmicx}  
\usepackage{algpseudocode}  

\begin{document}

\title{Jitter: Random Jittering Loss Function
\thanks{This work was supported by NJU Horizon Embedded Intelligence Fund.\\
* Corresponding author: Chenglei Peng (pcl@nju.edu.cn)}
}

\author{\IEEEauthorblockN{Zhicheng Cai,
Chenglei Peng\IEEEauthorrefmark{1} and Sidan Du}
\IEEEauthorblockA{\textit{School of Electronic Science and Engineering}\\
\textit{Nanjing University},
Nanjing, China 210023\\ Email: 181180002@smail.nju.edu.cn, \{pcl, coff128\}@nju.edu.cn}
}

\maketitle

\begin{abstract}
   Regularization plays a vital role in machine learning optimization. One novel regularization method called flooding makes the training loss fluctuate around the flooding level. It intends to make the model continue to ``random walk" until it comes to a flat loss landscape to enhance generalization. However, the hyper-parameter flooding level of the flooding method fails to be selected properly and uniformly. We propose a novel method called Jitter to improve it. Jitter is essentially a kind of random loss function. Before training, we randomly sample the ``Jitter Point" from a specific probability distribution. The flooding level should be replaced by Jitter point to obtain a new target function and train the model accordingly. As Jitter point acting as a random factor, we actually add some randomness to the loss function, which is consistent with the fact that there exists innumerable random behaviors in the learning process of the machine learning model and is supposed to make the model more robust. In addition, Jitter performs ``random walk" randomly which divides the loss curve into small intervals and then flipping them over, ideally making the loss curve much flatter and enhancing generalization ability. Moreover, Jitter can be a domain-, task-, and model-independent regularization method and train the model effectively after the training error reduces to zero. Our experimental results show that Jitter method can improve model performance more significantly than the previous flooding method and make the test loss curve descend twice.
\end{abstract}

\section{Introduction}
Machine learning is mainly challenged to make the trained model have excellent generalization performance. The model must perform well not only on the training set but also on new input that is not observed \cite{goodfellow2016deep}. Generally, when a machine learning model is trained, a loss function is set to measure and act as the target to reduce the training loss. The goal of the optimization method mainly focuses on minimizing generalization errors, also known as test errors. Overfitting and underfitting are two common reasons for the poor generalization ability of model, both of which are the result of the mismatch between the learning ability of the model and the data complexity. Overfitting is one of the most significant points of interest and concern in the machine learning community \cite{ng1997preventing,moore2001cross,caruana2001overfitting,belkin2018overfitting,roelofs2019meta}. Simply, overfitting refers to the phenomenon that the model is too complicated. The characteristics in the training set that do not apply to the test set are memorized, resulting in the huge gap between the training loss and the test loss. Whether the model is overfitted can be judged by observing whether the generalization gap obtained by test loss minus training loss keeps increasing or not.

To alleviate the overfitting problem and reduce the generalize error, many regularization methods have been proposed. For example, the penalty term of L2 norm constraint is added to the loss function to limit the weight decay with smaller L2 norm \cite{tikhonov2013numerical}. Other regularization methods with penalty terms have been proposed, such as elastic net regularization \cite{zou2005regularization} utilizing both L1 and L2 norm constraint \cite{tibshirani1996regression,tibshirani2005sparsity,yuan2006model}. Random inactivation of neurons forces individual neurons to learn more robust characteristics \cite{srivastava2014dropout}. Moreover, \cite{szegedy2016rethinking} combined the lower parameter count and additional regularization with batch-normalized auxiliary classifiers and Label-Smoothing regularization method. Data augmentation methods \cite{shorten2019survey,perez2017effectiveness,mikolajczyk2018data} add noise to the input, or simply stopping the training at an earlier phase when the validation accuracy does not ascend \cite{morgan1989generalization}. These regularization methods are considered to be able to directly control the training loss, reduce the generalization error, and enhance the generalization performance of the model to a certain extent.

\begin{figure}[htbp]
\centering
\includegraphics[height=0.25\textwidth]{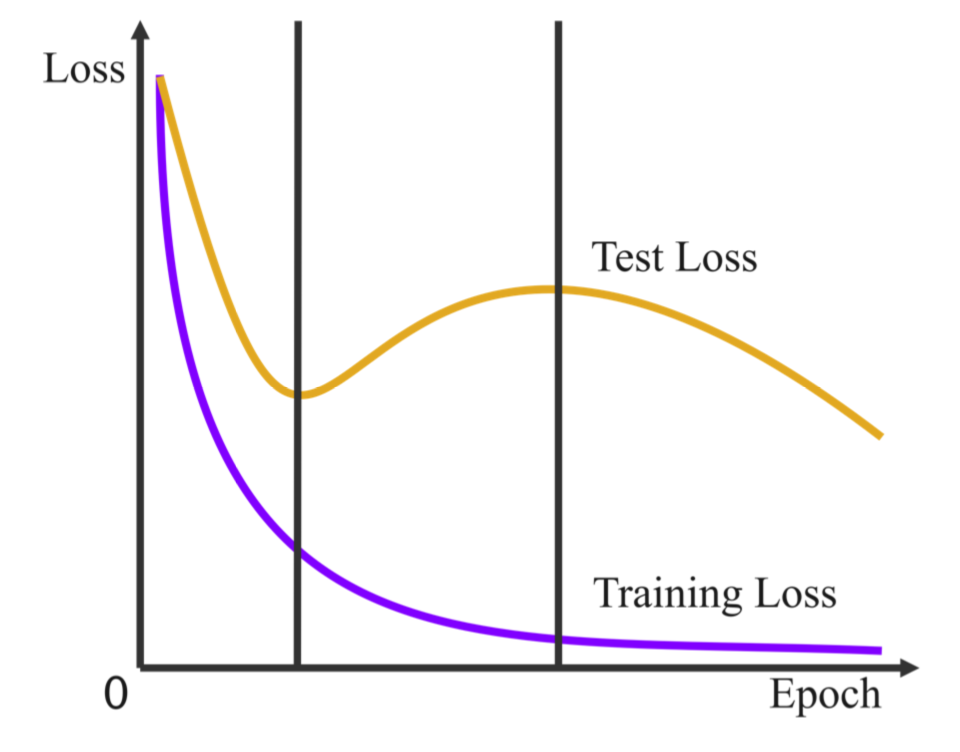}
\caption{Double descent curve phenomenon.}
\label{f1}
\end{figure}

Although stronger regularization makes training error and loss more difficult to approach zero, it can not keep the training loss at a correct level before the end of the training \cite{ishida2020we}. For some over-parametrized deep networks, the weak regularization method does not prevent training loss from dropping to zero, making it even harder to choose the hyper-parameter that corresponds to a specific loss level. \cite{ishida2020we} proposed a new regularization method under construction called flooding. Assuming that the original learning target is $J$, the improved flooding goal $\tilde{J}$ was shown in Eq.\ref{eq1}. 
\begin{equation}
\tilde{J}(\theta)=|J(\theta)-b|+b
\label{eq1}
\end{equation}

Where $b > 0$ is the flooding level specified by the user, which can be regarded as a hyper-parameter, and $\theta$ is the parameter of the model. The flooding is a direct solution to prevent the training loss from falling to zero when the training loss is reduced to the flooding level by random walking. As a byproduct, flooding method makes test loss curve appear double descent phenomenon \cite{nakkiran2019deep} (see Fig.~\ref{f1}).

However, the flooding level is obtained by performing the exhaustive search or observing the loss curve of the validation set. The exhaustive searching will lead to a large amount of time and computational cost. Moreover, it is often difficult to find the optimal value. In addition, the method of observing the loss curve of the validation set is not proper and rigorous. What's more, the optimal value can be changeable during the process of learning. 

Our paper proposes a novel method called Jitter, essentially a random loss function to improve the flooding method. In each back-propagation of the random jittering loss function during training, the $b$ in Eq.\ref{eq1} is randomly sampled from uniform distribution or Gaussian distribution, the value obtained is called Jitter point. To be more general, this paper also studies a method for obtaining the Jitter point value by random sampling from a standard normal distribution. Since the Jitter point value is obtained by random sampling, the loss function is therefore uncertain and changeable, which is  purposed to make the trained model possess better generalization ability. The experimental results show that the utilization of Jitter method can make the original model achieve better performance than the previous flooding method. As a universal and adaptable method, Jitter is domain- task- and model-independent. As a byproduct, the random jittering loss function can make the test loss curve descend twice.

What's more, it is known to us that there exists innumerate random behaviors during the process of the machine learning. However, the traditional target functions or loss functions of machine learning are all certain and unchangeable. This paper introduces randomness to the loss function first. This kind of random jittering target function achieved by the Jitter method is supposed to make the model more robust and possesses better generalization ability.

\section{Related Work}

\subsection{Regularization method}
Ivanov in 1962 first proposed the method of stable solution by adding constraints, where the basic idea was to exploit the energy of constrained restoration image \cite{ivanov1962linear}. In 1963, Tikhonov proposed the method of solving ill-conditioned problems and applied it to image restoration, which was called ``regularization" \cite{tikhonov1963solution,tikhonov1963regularization}. The basic idea of Tikhonov was to restrict the energy of high-frequency components in the restored image instead of the energy of the restored image. 

Nowadays, regularization is one of the core problems in machine learning realm. According to Goodfellow \cite{goodfellow2016deep}, regularization refers to modifying the learning algorithm to reduce the generalization error rather than the training error. As known from that, ``regularization" has further evolved to a more general meaning, including various methods that alleviate overfitting. There are more and more different kinds of regulation methods, according to the main idea of different regularization methods, the common regularization methods can be divided into \cite{khan2018guide}:

\begin{itemize}
\item Methods of regularizing models by data level technology (e.g., Data Augmentation \cite{2019Adversarial,2019Data,2021Data}, Label-Smoothing \cite{he2019bag,2020Delving,2020Towards});
\item Methods of introducing random behavior into neuronal activation (e.g., Drop Block \cite{ghiasi2018dropblock}, AutoDropout \cite{2021AutoDropout});
\item Method of normalizing batch statistics in feature activation (e.g., Batch Group Normalization \cite{2020Batch}, SelfNorm and CrossNorm \cite{2021SelfNorm});
\item Avoid overfitting by decision level fusion (e.g., Bagging \cite{breiman1996bagging}, Boosting \cite{schapire1999brief,2016xgboost});
\item Methods of introducing norm constraints on network weights (e.g., L2 norm constraint, elastic network constraint);
\item Utilize guidance from validation set to control the learning process (e.g., Early Stopping).
\end{itemize}

Our Jitter method actually regularizes the model from the aspect of loss function.

\subsection{The phenomenon of double descent curves}
The phenomenon of ``double descent" was first discovered by Krogh and Hertz \cite{krogh1991simple} in 1992, where they showed the double descent phenomenon under a linear regression setup theoretically. But this phenomenon has not been named until 2019 by Belkin \cite{belkin2019reconciling} to explain the two stages of deep learning. In the first stage, where the model complexity is small compared to the number of training samples, that is to say, the model is underfitting. The test error curve decreases with the increase of model complexity. When the model complexity increases to a certain extent, the test error curve begins to rise. It confirms the view of traditional machine learning that too much model complexity will reduce model generalization ability. In the stage II, the model is over-parameterized, which means the model complexity becomes even more extensive. Then the curve starts to descend again as increasing the complexity decreases test error, leading to the formation of double descent shape. The phenomenon that the test error decreases again usually occurs after the training error falls to zero. It is consistent with the view of modern machine learning that larger models have better generalization performance.  In our paper, the double descent curve phenomenon is observed in the evolving process of test loss.

\subsection{Flooding method}

In this subsection, we discuss the regularization method of flooding. The idea of flooding method is ``floods the bottom area and sinks the original empirical risk, so that the essential risk cannot go below the flooding level" \cite{ishida2020we} . 
As is shown in Eq.\ref{eq1}, if $J(\theta) > b$, then $\tilde{J} (\theta) $ and $J(\theta) $ have the same phase; if $J (\theta) < b$, then the phase of  $\tilde{J}(\theta) $ is opposite to that of $J(\theta) $. It means that if the initial learning goal is above the flooding level, the gradient will descend, and the gradient ascends otherwise. The flooding method makes the training loss increase and decrease in the region near the flooding level, 
making the model random walk in the area of non-zero training loss. The intention is to make the model enter a flat loss area, make the loss curve flatter, and finally improve the generalization performance of the model.

As mentioned above, flooding level can be regarded as a user-specified hyper-parameter, and the better value can be obtained through exhaustive search or validation set guidance. However, we considers that the two methods are time-consuming, laborious and not rigorous. Besides, there exists no unified and standard configurations for different models and tasks. Moreover, with the increase of training rounds, the better level is likely to change. That is to say, setting the flooding level as a constant can make the training loss fluctuate up and down in a specific range, however, as for different training periods, the most appropriate range which is time-depend may be diverse.


\section{Random Jittering Loss Function - Jitter}

\subsection{Jitter algorithm}

According to the previous section analysis, the flooding method has certain innovations and advantages, but some shortcomings also exist. Therefore, this paper proposes Jitter as an improvement of flooding. In each forward propagation of the model, Jitter randomly samples a value $\alpha $ from a specific probability distribution  within a certain interval like the uniform distribution or Gaussian distribution. The resulting $\alpha $ is called Jitter point. The expression of random jittering loss function is shown in Eq.\ref{eq2}.

\begin{equation}
\begin{split}
\tilde{L}(\theta) =|L(\theta)&-\alpha|+\alpha  \\ 
\alpha \sim \mathcal N \left (\mu,\sigma^2 \right) \ & or \  \alpha \sim U \left (a,b \right)
\label{eq2}
\end{split}
\end{equation}
Where, $L(\theta)$ is the initial loss function, $\tilde{L}(\theta)$ is the improved random jittering loss function with Jitter point, $\alpha$ is the Jitter point which is randomly sampled from a specific distribution, and $\theta$ stands for the parameters of the model.

When $L(\theta) > \alpha$, $\tilde{L}(\theta)$ has the same form as $L(\theta)$, and the initial learning target is affected by the ``gravity" effect , leading to gradient descending during backpropagation. On the contrary, when $L(\theta) < \alpha$, $\tilde{L}(\theta)$ is on the opposite side of $L(\theta)$, the initial learning target gets a ``buoyancy" effect, leading to the gradient ascending in the opposite direction. Moreover, the Jitter method is a generic regularization method, and it is domain-, task-, model-independent. The Jitter method will also cause the phenomenon of double descent test loss curve, so that when the training error is reduced to zero, the model can still carry out effective training.

\subsection{Implementation}
Mini-batched stochastic optimization which makes computation efficient is often utilized for large scale problems. Suppose that $D=\{(\mathbf{X}_i,\mathbf{y}_i)\}^N_{i=1}$ represents a dataset with $N$ samples, $M$ stands for the number of disjoint mini-batches, and the batch size is $N_b$. Here we summarized the pseudo code of Jitter in Algorithm ~\ref{alg1}.

\renewcommand{\algorithmicrequire}{\textbf{Input:}}
\renewcommand{\algorithmicensure}{\textbf{Output:}}

\begin{algorithm}[H]	
	\caption{The pseudo code of Jitter}
	\label{alg1}
	\begin{algorithmic}[1]
      \Require 
         \Statex $D=\{(\mathbf{X}_i,\mathbf{y}_i)\}^N_{i=1}$: Training dataset,
         \Statex $N_b$: Batch size,
         \Statex $M$: Number of disjoint mini-batches,
         \Statex $\mathcal D(a,b)$: Specific distribution where Jitter point sampled from
      \Ensure
         \Statex $\mathbf{y}_p$: Predicted label vector
      \Repeat
         \For{t=1: $M$}
            \State Randomly select $N_b$ instances from $D$
            \State Sample Jitter point $\alpha_m$ from $\mathcal D(a,b)$ randomly
            \State Update the loss function with Eq.\ref{eq2}
            \State Input $\mathbf{X}^t$ to the model
            \State Compute the label prediction vector $\mathbf{y}_p^t$
            \State Calculate the loss with Eq.\ref{eq2}
            \State Update model parameters by back-propagation
         \EndFor
      \Until{converge}\\
      \Return $\mathbf{y}_p$
   \end{algorithmic}  
\end{algorithm}

In addition, suppose that $\tilde{R}(\mathbf{X})$ stands for the empirical risk of full-batch case, and ${R_m}(\mathbf{X}_m)$ represents the original empirical risk of the m-th mini-batch for $m \in \{1,...,M\}$. According to Eq.\ref{eq2}, the actually empirical risk of the m-th mini-batch with Jitter point $\alpha_m$ is $\tilde{R_m}(\mathbf{X}_m)= |R(\mathbf{X}_M)-\alpha_m|+\alpha_m$. On the basis of the convexity of the absolute value function and Jensen's inequality, we can prove that by mini-batched stochastic gradient descent with Jitter can minimize the upper bound of the full-batch case empirical risk. That is:
\begin{equation}
\tilde{R}(\mathbf{X}) \leq \frac{1}{M}\sum^M_{m=1}\left(|R_m(\mathbf{X}_m)-\alpha_m|+\alpha_m \right)
\label{eq3}
\end{equation}

\begin{figure*}[htbp]
\begin{center}
\subfigure[Original method]{
\includegraphics[width=0.3\textwidth]{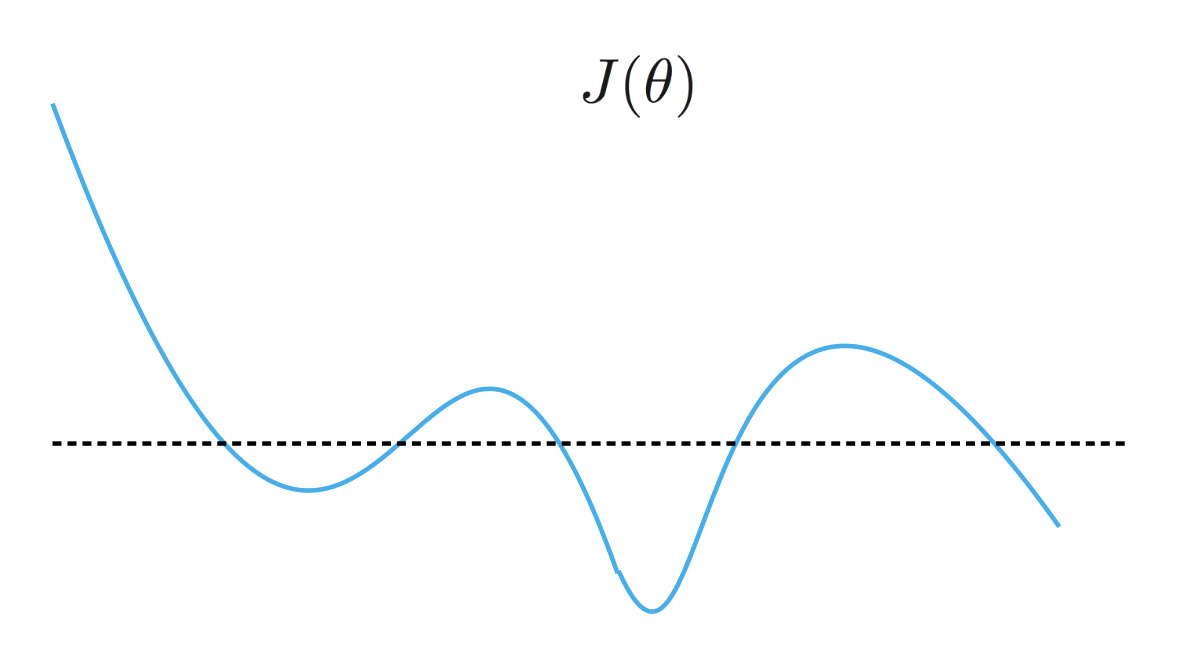}}
\subfigure[Flooding method]{
\includegraphics[width=0.3\textwidth]{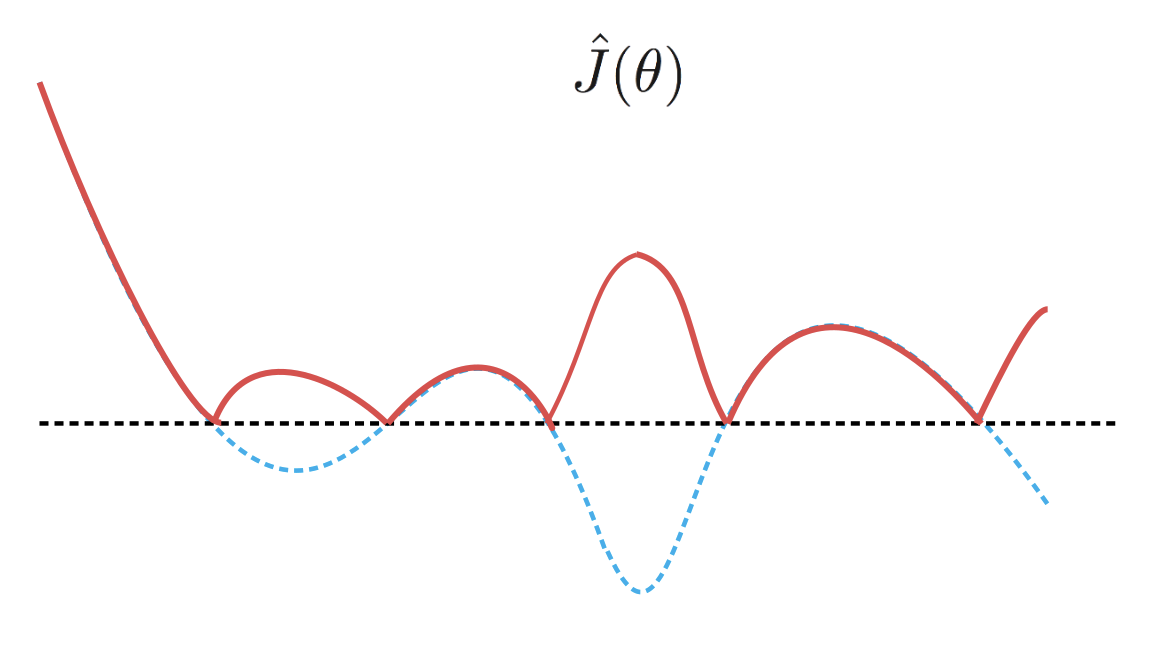}}
\subfigure[Jitter method]{
\includegraphics[width=0.3\textwidth]{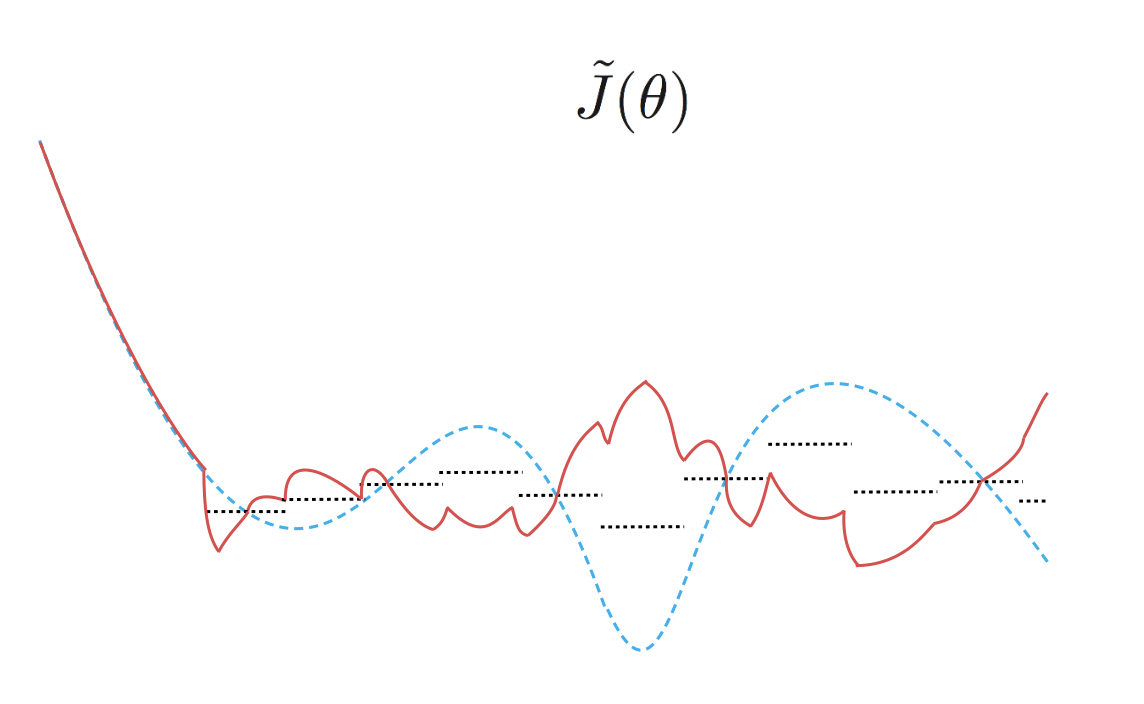}}
\end{center}
\caption{Schematic diagram of loss functions of original method (a), flooding method (b), and Jitter method (c).}
\label{f2}
\end{figure*}

\subsection{Advantages of Jitter method}
In this section, we talk about the specific advantages which Jitter possesses.

Because the Jitter point is obtained by randomly sampling, it does not need to be selected exhaustively as the flooding level value. Our experimental results have validated that sampling Jitter point from a standard normal distribution with a average value of 0 and a standard deviation of 1 can still obtain good results. This random sampling method named as Standard Jitter method provides a unified standard for parameter selection, as a result, it becomes unnecessary to select the hyper-parameter by detecting the loss curve of the validation set.

In addition, unlike the flooding method, which uses a fixed constant flooding level, since Jitter point is obtained by sampling randomly, our loss function is also random. In this way, adding randomness tries to make the training model be able to resist some random and unknown input data changes, enhance the robustness of the learning features of the model, and improve the generalization ability of the model further. This design is actually consistent to the inevitable randomness in the learning procedure of the machine learning models. 

Moreover, Jitter breaks through the previous method which makes the training loss curve fluctuate up and down around the certain level in the late training period. The random loss function makes the training loss fluctuate within a certain interval according to the probability, which makes the ``random walk" of the model occur randomly. Therefore, the loss curve is divided into cells for flipping, which has a greater probability of resisting the uneven part of the loss curve, making the loss curve flatter than the flooding method and enhancing the model's generalization performance more significantly. As shown in Fig.~\ref{f2}, the loss functions of the original method without flooding or Jitter method, flooding method and Jitter method are shown respectively.

Here, we take the one dimension loss function curve as an example. Using the flooding method is equivalent to turning up the part below the flooding level threshold. It can be found that there are many more local minimum values for the whole target. Moreover, the number of local minimum values will grow exponentially when the dimension of the loss function curve increases. We can consider the loss of flooding $\hat{J}(\theta) $ is flatter than the original loss of ${J}(\theta) $, so the generalization ability of the model will be better. Since the loss function is flipped in multiple intervals, the Jitter method intuitively make the loss flatter compared with the flooding method.

\subsection{Theoretical analysis}

A simple explanation for Jitter method is that Jitter method increases the number of local minimum values of the original loss function, and equips the model with certain climbing ability, which can prevent the optimization of parameters from falling into a bad local minimum value and unable to jump out.

\begin{figure}[htbp]
\begin{center}
\subfigure[The steeper loss function]{
\includegraphics[width=0.22\textwidth]{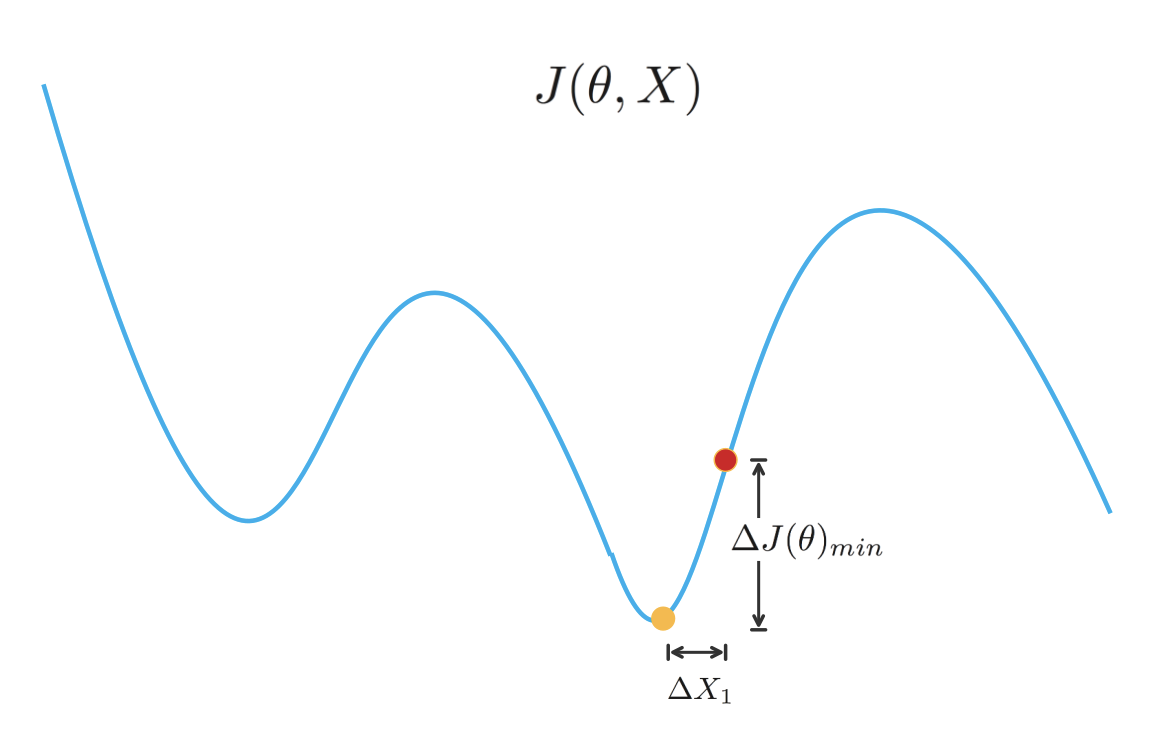}}
\subfigure[The flatter loss function]{
\includegraphics[width=0.22\textwidth]{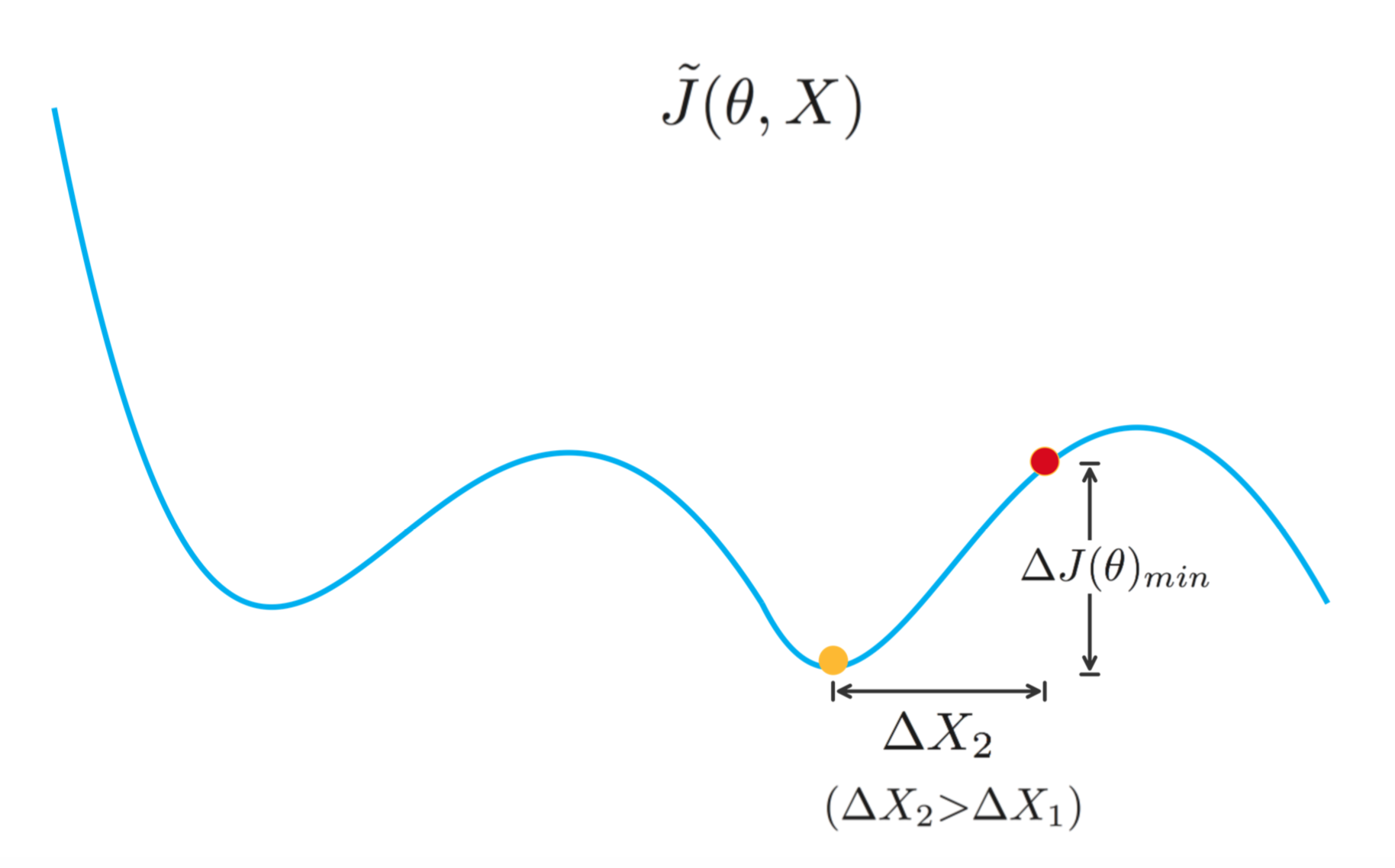}}
\end{center}
\caption{Schematic diagram of two loss functions corresponding to normal sample and adversarial sample.}
\label{f3}
\end{figure}

The analysis is then performed from the point of view of the adversarial sample. The adversarial sample makes the loss larger by generating wrong samples, so that the model classifies the generated sample incorrectly. We assume that the least loss difference of the model between the correct and incorrect classification is $\triangle J(\theta)_{min} $, that is, when the loss of the counter sample is higher than the loss difference of $\triangle J(\theta)_{min} $, the adversarial sample will be misclassified. Intuitively, the flatter the loss is, the farther the interval between the counter sample and the normal sample is, and the more difficult it is to generate it. Therefore, the flatter the loss is, the more robust the model will be to counter disturbance. Fig.~\ref{f3} is the schematic diagram of two loss functions corresponding to normal sample and adversarial sample. The orange dot represents the normal paired sample, the red dot represents the adversarial sample, and the right graph loss function is flatter than the left. Similarly, the steeper the loss function is, the larger the interval between the adversary sample and the normal sample is, and the more steep the loss function is, the more likely the adversary sample will be misclassified.
   
In fact, the general robustness and generalization are also the same. The general robustness refers to the robustness of the model to some samples, such as Gaussian fuzzy, salt-and-pepper noise. In other words, if the sample is perturbed to a certain extent, the loss change of the model to the disturbed sample should not be too large. As a result, the flatter the loss, the better the robustness. In other words, for an unknown sample and a labeled sample belonging to the same class, the necessary condition for the model to divide it into pairs is that the loss difference cannot be too large, then the ``flat" loss can meet this condition, thus the generalization will be better.

From the perspective of SVM \cite{noble2006support}, for a linear separable binary classification problem, there are innumerable classification super-planes to separate it. SVM is to select the classifier that can meet the ``maximum interval". From another point of view, the smoother the loss, the more likely it is to separate different classes. Because if the sample is slightly disturbed, the change of loss will not be too large, which means that the sample with slight disturbance will not run to the other side of the classification surface.

\subsection{Mathematical analysis}

\textbf{Theorem I: the equivalent effective flooding value of Jitter point value sampled by Standard Jitter method is $\frac {1} {\sqrt {2\pi}}$.}
\\

It is easy to know that the equivalent theoretical flooding value of Jitter point value is the expectation value of the distribution that the Jitter point sampled from. So, in terms of the standard Jitter method which samples from the standard normal distribution whose expectation is zero, the equivalent theoretical flooding value is zero.

However, in the flooding method, the value of flooding level is positive. Also, the value of the loss function $L(\theta)$ is constant positive as well. As a result, the Jitter points that make an effect actually are these positive ones, which means that the negative Jitter points make no difference. So equivalent effective flooding value can be inferred as below.

The probability density function of standard normal function is:
\begin{equation}
   \begin{split}
   f(x)&=\frac{1}{\sigma\sqrt{2\pi}}e^{-\frac{(x-\mu)^2}{2\sigma^2}}\\
       &=\frac{1}{\sqrt{2\pi}}e^{-\frac{x^2}{2}}
   \label{eq4}
   \end{split}
\end{equation}

The expectation value of effective Jitter points $\alpha$ is:
\begin{equation}
   \begin{split}
   E(\alpha)&=\int^{\infty}_{-\infty} f(\alpha)\alpha\cdot d\alpha\\
            &=\int^{\infty}_{0} \frac{1}{\sqrt{2\pi}}e^{-\frac{\alpha^2}{2}}\alpha\cdot d\alpha\\
            &=\frac{1}{\sqrt{2\pi}}\int^{\infty}_{0} e^{-\frac{\alpha^2}{2}}\cdot d\frac{\alpha^2}{2}\\
            &=\frac{1}{\sqrt{2\pi}}\int^{\infty}_{0} e^{-x}\cdot d x\\
            &=\frac{1}{\sqrt{2\pi}}
   \label{eq5}
   \end{split}
\end{equation}

Now, we have proved that equivalent effective flooding value is $\frac {1} {\sqrt {2\pi}}$.
\\

\textbf{Theorem II: Jitter method increases the expected value of training loss, but reduces the mean square error of the estimated empirical risk.}

If the Jitter point $\alpha$ satisfies $L(\theta) \geq \alpha$, we have:
\begin{equation}
   \begin{split}
   \tilde {L}(\theta) &= |L(\theta)-\alpha|+\alpha\\
                    &= L(\theta)
   \label{eq6}
   \end{split}
\end{equation}

If the Jitter point $\alpha$ satisfies $L(\theta) < \alpha$, we have:
\begin{equation}
   \begin{split}
   \tilde{L}(\theta) &= |L(\theta)-\alpha|+\alpha\\
                    &= 2\alpha-L(\theta)\\
                    & > L(\theta)
   \label{eq7}
   \end{split}
\end{equation}

So, $L(\theta) \leq \tilde{L}(\theta)$ is established, as a result, the expectation value of $\tilde{L}(\theta)$ is not less than that of $L(\theta)$.

However, the mean squared error(MSE) of the estimated empirical risk is smaller than the original risk estimator.
If the Jitter point $\alpha$ satisfies $\hat{L}(f) < \alpha < L(f)$, we have:
\begin{equation}
   \begin{split}
   MSE(\tilde{L}(f)) < MSE(\hat{L}(f))
   \label{eq8}
   \end{split}
\end{equation}

And if the Jitter point $\alpha$ satisfies $\alpha \leq \hat{L}(f)$, we have:
\begin{equation}
   \begin{split}
   MSE(\hat{L}(f))=MSE(\tilde{L}(f))
   \label{eq9}
   \end{split}
\end{equation}

Here, $f$ stands for any measurable vector-valued function. $\hat{L}(f)$ is the original training loss, $\tilde{L}(f)$ is the actual training loss and $L(f)$ is the test loss. The proof is similar with that in \cite{ishida2020we}, but replacing the flooding level $b$ with the expectation value of Jitter point $\alpha$. 

\section{Experiment}

\subsection{Dataset description }

We tested various methods on six benchmark datasets: CIFAR-10, CIFAR-100, SVHN, MNIST, KMNIST and Fashion MNIST. 
\begin{itemize}
\item CIFAR-10: CIFAR-10 dataset is composed of 10 classes of natural images with 50,000 training images and 10,000 testing images in total. The 10 classes include: airplane, automobile, bird, cat, deer, dog, frog, horse, ship and truck. Each image is a RGB image of size $32\times32$. 
\item CIFAR-100: CIFAR-100 dataset is composed of 100 different classifications, and each classification includes 600 different color images, of which 500 are training images and 100 are test images. As a matter of fact, these 100 classes are composed of 20 super classes, and each super class possesses 5 child classes. The images in the CIFAR-100 dataset have a size of $32\times32$ like CIFAR-10. 
\item SVHN: SVHN (Street View House Numbers) dataset is composed of 630,420 RGB digital images with a size of $32\times32$, including a training set with 73,257 images and a test set with 26,032 images. 
\item MNIST: MNIST is a 10 class dataset of 0 $\sim$ 9 handwritten digits,including 60,000 training images and 10,000 test images in total. Each sample is a size $28\times28$ gray-scale image. 
\item KMNIST: Kuzushiji-MNIST (KMNIST) dataset is composed of 10 classes of cursive Japanese characters (namely, ``Kuzushiji").  Each sample is a gray-scale image of $28\times28$ size. This dataset includes 60, 000 training images and 10,000 test images in total. 
\item Fashion-MNIST: Fashion-MNIST (FMNIST for short) is a 10 class dataset of fashion items:T-shirt/top, Trouser, Pullover, Dress, Coat, Scandal, Shirt, Sneaker, Bag and Ankle boot. Each sample is a gray-scale image of $28\times28$ size. This dataset includes 60,000 training images and 10,000 test images in total.
\end{itemize}

\begin{table*}[ht]
  \caption{The configurations of six kinds of Jitter methods.}
   \label{tab1}
  \centering
   \begin{tabular}{|l|c|c|c|c|}
   \hline
   \textbf{Method} & \textbf{Distribution} & \textbf{Interval} & \textbf{Average Value} & \textbf{Standard Deviation}  \\
   \hline
   Jitter\_1 & Uniform Distribution & [0.00,0.04] & 0.02 & $\sqrt{0.04/3}$ \\
   \hline
   Jitter\_2 & Uniform Distribution & [0.01,0.03] & 0.02 &  $\sqrt{0.02/3}$ \\
   \hline
   Jitter\_3 & Gaussian Distribution & [0.00,0.04] & 0.02 & 0.01 \\
   \hline
   Jitter\_4 & Gaussian Distribution & [0.01,0.03] & 0.02 & 0.005 \\
   \hline
   Jitter\_5 & Normal Distribution & [$-\infty$,$+\infty$] & 0 & 0.1 \\
   \hline
   Jitter\_S & Normal Distribution & [$-\infty$,$+\infty$] & 0 & 1 \\
   \hline
   \end{tabular}
\end{table*}

\begin{table*}[ht]
   \caption{Test accuracy of various methods on different benchmark datasets.}
    \label{tab2}
   \centering
    \begin{tabular}{|c|c|c|c|c|c|c|}
    \hline
    \textbf{Method} & \textbf{CIFAR-10} & \textbf{CIFAR-100} & \textbf{SVHN} & \textbf{MNIST} & \textbf{KMNIST} & \textbf{F-MNIST} \\
    \hline
    Accuracy(\%) & best \ \ \ mean & best\ \ \  mean & best\ \ \  mean & best\ \ \  mean & best\ \ \  mean & best\ \ \  mean \\
    \hline
    Original & 90.35\ \ 90.34 & 65.85\ \ 65.82 &92.15\ \ 92.13 &99.23\ \ 99.23 &96.08\ \ 96.05    &92.46\ \ 92.43 \\
    \hline
    Flooding & 90.49\ \  90.47  &66.13\ \ 66.12 &93.14\ \ 93.13 &99.26\ \ 99.25 &96.66\ \ 96.63 &93.29\ \ 93.26\\
    \hline
    Jitter\_1 & 90.48\ \ 90.45  &66.30\ \ 66.28 &93.09\ \ 93.03 &99.37\ \ 99.36 &96.72\ \ 96.70 &92.97\ \ 92.95\\
    \hline
    Jitter\_2 & 90.58\ \ 90.56 &66.43\ \ 66.41 &93.20\ \ 93.16 &99.33\ \ 99.33 &96.67\ \ 96.62  &93.24\ \ 93.22\\
    \hline
    Jitter\_3 & 90.55\ \ 90.53 &65.93\ \ 65.85 &93.10\ \ 93.07  &99.30\ \ 99.30 &96.64\ \ 96.63 &93.17\ \ 93.14\\
    \hline
    Jitter\_4 & 90.55\ \ 90.54 &65.93\ \ 65.88 &\textbf{93.29}\ \ \textbf{93.26} &99.37\ \ 99.36 &96.63\ \ 96.62 &92.65\ \ 92.64\\
    \hline
    Jitter\_5 & \textbf{90.61}\ \ \textbf{90.57} &\textbf{66.65}\ \ \textbf{66.63} &93.01\ \ 92.89 &\textbf{99.40}\ \ \textbf{99.39} &\textbf{96.76}\ \ \textbf{96.75} &\textbf{93.33}\ \ \textbf{93.31}\\
    \hline
    Jitter\_S & 89.73\ \ 89.69  &65.54\ \ 65.47 &91.93\ \ 91.87 &99.04\ \ 99.03 &95.99\ \ 95.97  &92.24\ \ 92.19\\
    \hline
    \end{tabular}
 \end{table*}

\begin{figure*}[ht]
   \centering
   \subfigure[Original method]{
   \includegraphics[width=0.38\textwidth]{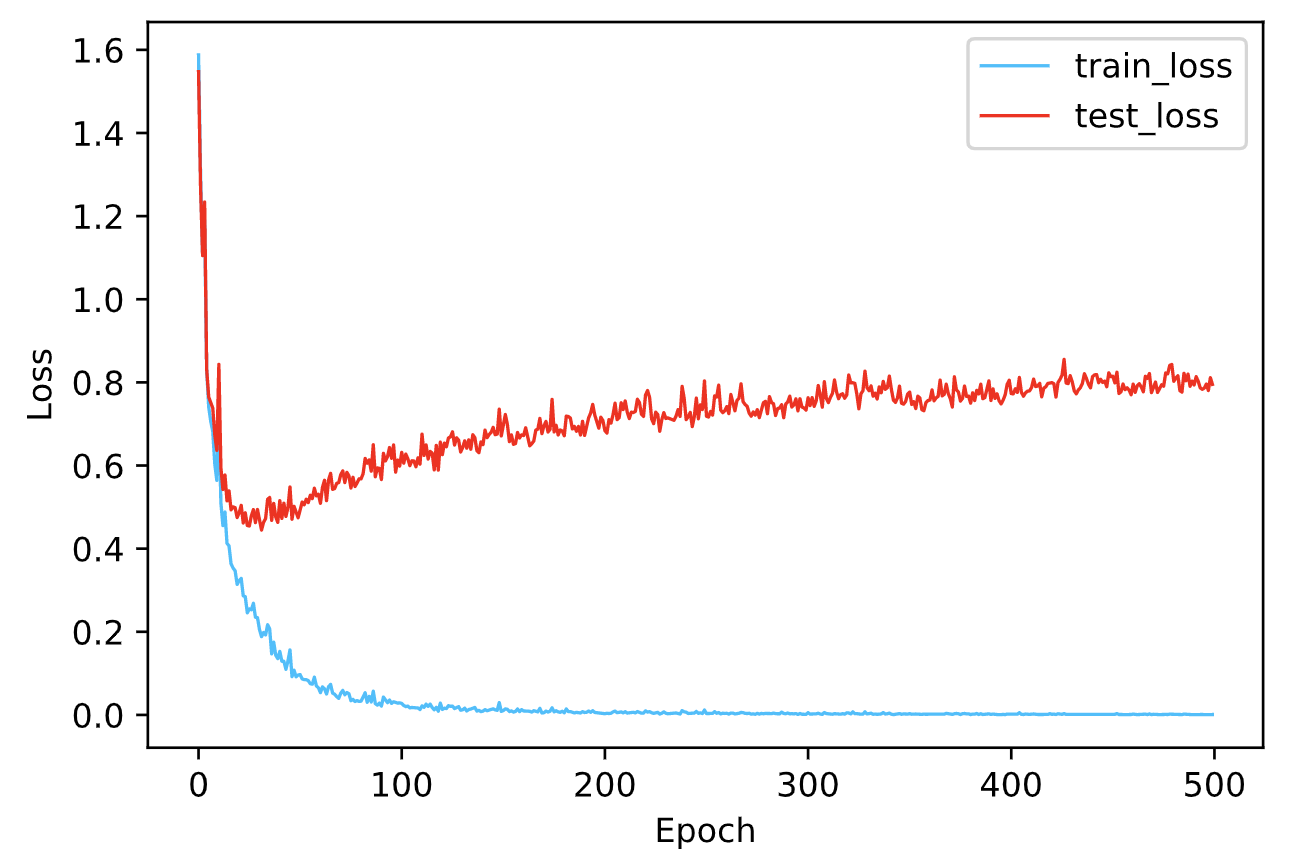}
   }
   \quad
   \subfigure[Jitter\_1 method]{
   \includegraphics[width=0.38\textwidth]{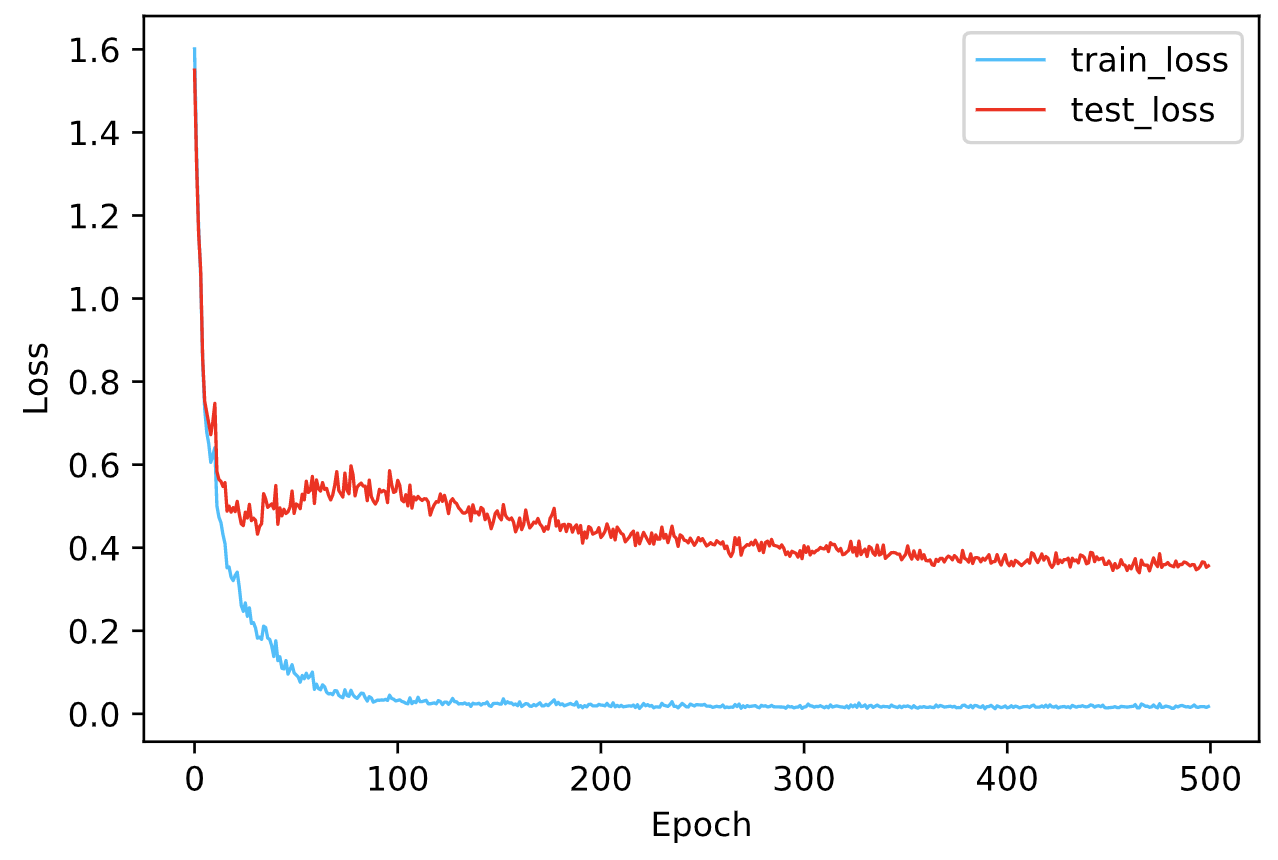}
   }
   \quad
   \subfigure[Jitter\_5 method]{
   \includegraphics[width=0.38\textwidth]{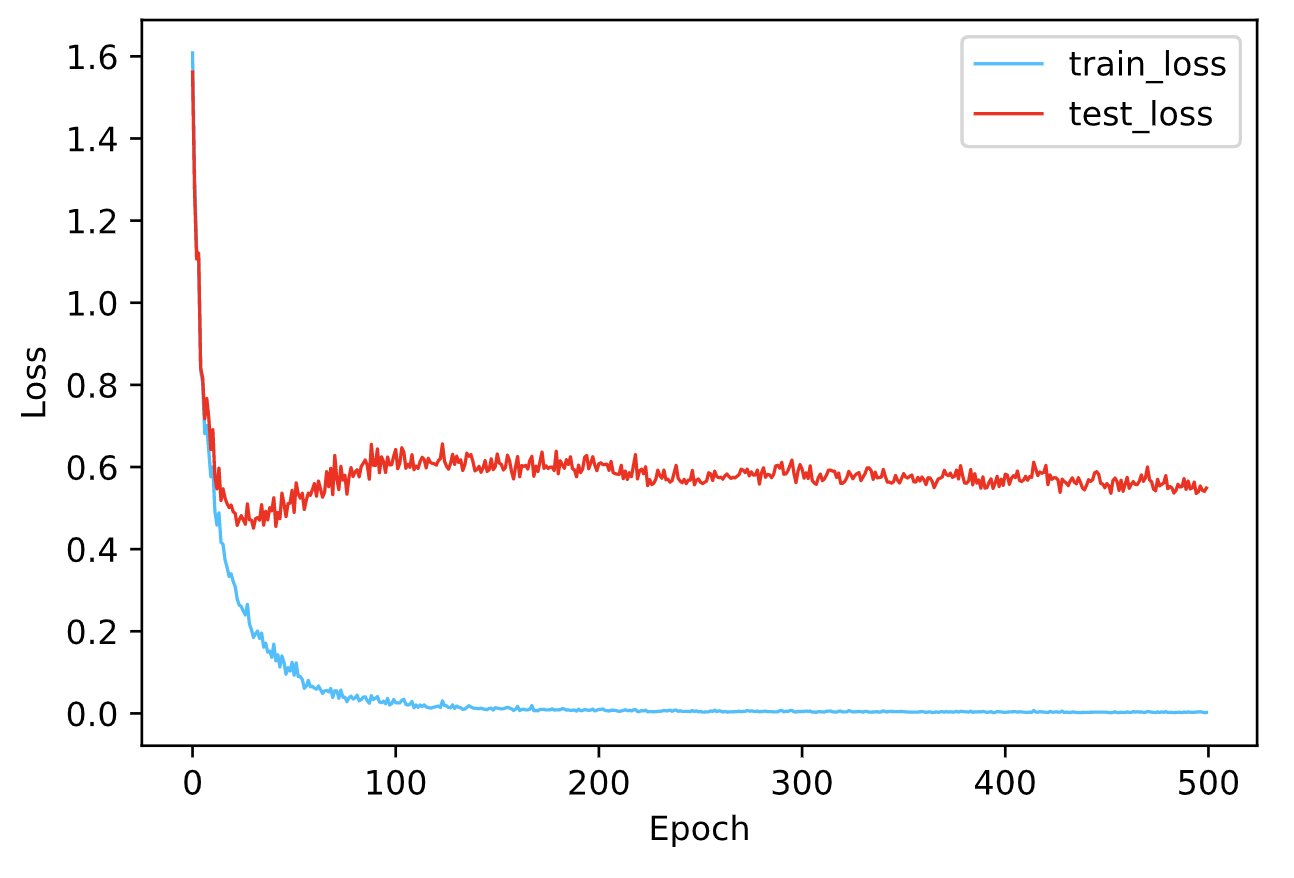}
   }
   \quad
   \subfigure[Jitter\_S method]{
   \includegraphics[width=0.38\textwidth]{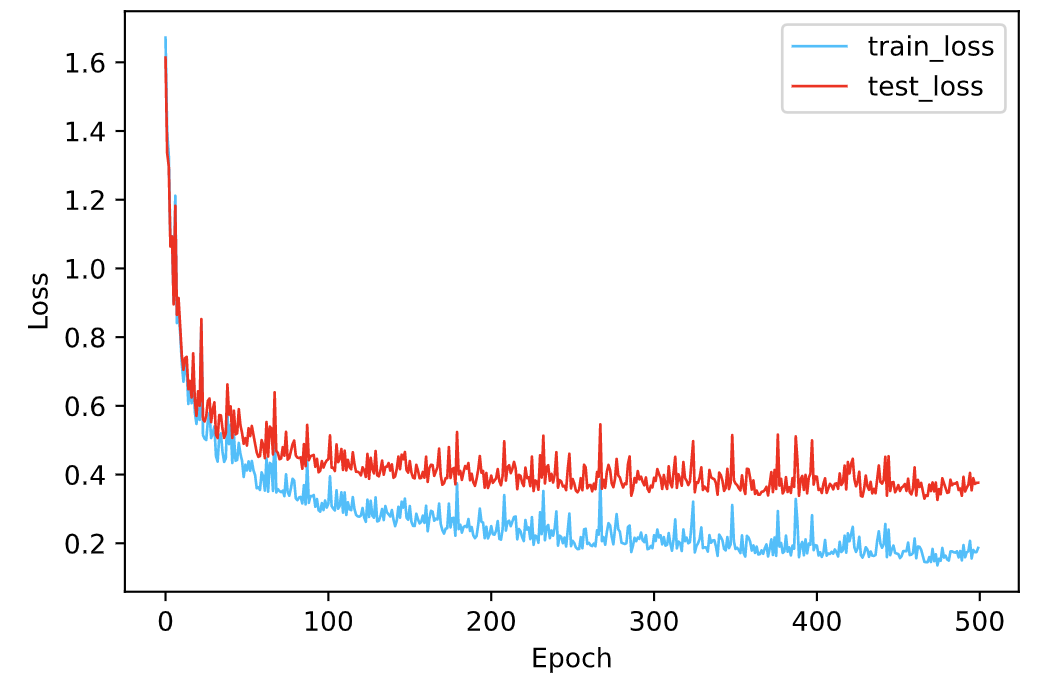}
   }
   \caption{Training loss and test loss curves of various methods on CIFAR-10.}
   \label{f4}
\end{figure*}

\subsection{Basic configuration}

In this paper, the convolutional neural network model  VGG-16\cite{simonyan2014very} is selected as the basic model, which is composed of 13 layers of convolution layer and 3 layers of fully connected layer. The convolution layers with step-size is utilized to conduct downsampling operation instead of employing pooling method~\cite{cai2021study}. As a regularization method, we add the dropout to the first two of the three fully connected layers and set the activation rate to $50\%$. In addition, we add batch normalization layers after all convolutional layers. All the models use cross-entropy loss as the original loss function. The training algorithm uses mini-batch stochastic gradient descent with a momentum term of 0.95 and weight decay coefficient of 0.0005, the batch size is 128.  All experimental models have been trained with NVIDIA GeForce GTX 2080Ti GPU.

For CIFAR-10, CIFAR-100 and SVHN, the model is trained for 500 epochs. For others, it is 300 epochs. During the training, the learning rate remains unchanged with 0.001. For the training data of CIFAR-10 and CIFAR-100 data sets, we adopt the method of data augmentation as follows: four circles of zero pixels are padded around the original image, and then the padded image was randomly cropped to the size of the original image. Then we flip the image horizontally at a probability level of 0.5. 

\subsection{Compared methods}

Eight groups of comparative experiments are carried out on all the selected five data sets. The first experiment is original models without the Jitter method or flooding method. The second one is the flooding method with the flooding level of 0.02. The third one is the Jitter\_1 model with Jitter point sampled directly from a uniform distribution on an interval of [0.00, 0.04]. The fourth one is the Jitter\_2 model of Jitter point sampled directly from a uniform distribution on an interval of [0.01, 0.03]. The fifth one is the Jitter\_3 model of Jitter point sampled directly from Gaussian distribution on the interval [0.00,0.04] with the average value of 0.02 and standard deviation of 0.01. The sixth one is the Jitter\_4 model of Jitter point sampled directly from Gaussian distribution on the interval [0.01,0.03] with the average value of 0.02 and standard deviation of 0.005. Considering that the training loss of the model always has an order less than 1e-2, so the equivalent effective flooding value of the standard Jitter method, $\frac{1}{\sqrt{2\pi}}$ is slightly unsuitable for the original loss function. As a result, we multiply the Jitter point sampled from the standard normal distribution with a correction factor 1e-1. And this is the seventh method Jitter\_5. The last one is the Jitter\_Standard(Jitter\_S for short) model of Jitter point from the normal distribution. To be seen more clearly, the configurations of these six kinds of Jitter methods are shown in Table.~\ref{tab1}.

For the Jitter\_1$\sim$4 models, the Jitter point sampled has an expected value of 0.02, which is the same as the flooding level value selected in the flooding method, and the purpose of this setting is to compare the random Jitter method with the fixed method in which the flipping level of the lost function has the same expected value. In fact, however, for these four models, distributions of different kinds, such as uniform distribution, Gaussian distribution, and the choice of mean, variance or sampling interval of the distributions can still be considered artificial selection of hyperparameters. To provide a more uniform and straightforward standard selection of Jitter point values, we set up the Jitter\_S model that gets the Jitter point values sampled from a standard normal distribution. We find that this method makes the test loss curve similar to the training loss curve in that it keeps descending and converges finally, which means the overfitting problem of the model does not show up.

\subsection{Experimental results and analysis}

Table.~\ref{tab2} shows all the results of our experiments. We run each model five times. The variances were close to zero, therefore not listed in the Table.~\ref{tab2}. From Table.~\ref{tab2}, we can see that the methods Jitter\_1$\sim$5 and flooding all achieve better results than the original method, which indicates that these improved methods can indeed enhance the generalization ability of the original model. In addition, most of the  Jitter methods outperform the flooding method, validating the effectiveness of our Jitter method. The method Jitter\_5 achieves the best generalization results on most benchmark datasets which are represented by the test accuracy. To explore the reason, we think that the normal distribution possesses the maximum uncertainty and the minimum prior knowledge, which enhances the generalization ability. Besides, according to the central limit theorem, normal distribution is supposed to modeling the optimal Jitter point sampling. On the other side, the test accuracy of the method Jitter\_S is slightly lower than the original and other improved methods, we attribute this phenomenon to that the dispersion and volatility of the distribution utilized in the method Jitter\_S is excessive, which may be unsuitable for the loss curve. As a result, we argued that the correction factor employed in Jitter\_5 method is significant.

Fig.~\ref{f4} shows the training and test loss curves of the Original method (a), Jitter\_1 method (b), Jitter\_5 method (c) and the Jitter\_S method (d) on CIFAR-10 dataset. From Fig.~\ref{f4} (a),  We can see that the testing loss curve of the model trained in original method begins ascending in about 40-th epoch, indicating the emerging of over-fitting. Fig.~\ref{f4} (b) exhibits the double descent phenomenon of the Jitter\_1 method clearly. For the Jitter\_5 method, the test loss keeps unchanged virtually after the first ascent. Fig.~\ref{f4} (d) shows that the test loss curve of Jitter\_S keeps descending and converges finally like the training loss curve, which means that though the double descent phenomenon fails to arise, the model using Jitter\_S does not suffer the problem of over-fitting like the original method. The disappearance of the double descent phenomenon may contribute to the high uncertainty and volatility of the distribution utilized, which alternates the learning difficulty and optimization direction. Moreover, compared to the test loss of other methods, the method Jitter\_S obtains the lowest test loss but has the highest test error. This paradox may make us think about whether we should achieve a low test loss or a low test error and whether the training after achieving zero training error is meaningful.


\section{Conclusion}
In this paper, we proposed a novel regularization method called ``Jitter". Jitter method samples the value of ``Jitter Point" from a uniform distribution or a Gaussian distribution and introduces randomness to the loss function. Our experiments validate that the Jitter method can improve the original model’s generalization ability. In addition, Jitter method achieved better generalization results than flooding method on various benchmark data sets including CIFAR-10, CIFAR-100, SVHN, MNIST, KMNIST and FashionMNIST. Moreover, Jitter method can make the test loss curve descend for the second and converge finally, which means the model can be trained efficiently as well when the train error tends to zero. As a domain-, task-, and model-independent regularization method, Jitter method can be utilized in all the machine learning domains. 

Essentially, Jitter method is a kind of random loss function proposed first. There exist innumerable random behaviors in the training and inferring process of the machine learning models. We think that further research on random loss function might be beneficial to the improvement of machine learning.

\end{document}